\documentclass[10pt,letterpaper]{article}

\usepackage{amsmath,amsfonts,bm}

\def\eqref#1{equation~\ref{#1}}

\def\1{\bm{1}}

\def\vmu{{\bm{\mu}}}

\def\vs{{\bm{s}}}

\def\vu{{\bm{u}}}
\def\vv{{\bm{v}}}

\def\vx{{\bm{x}}}

\def\vz{{\bm{z}}}

\def\mJ{{\bm{J}}}

\def\mSigma{{\bm{\Sigma}}}

\DeclareMathAlphabet{\mathsfit}{\encodingdefault}{\sfdefault}{m}{sl}
\SetMathAlphabet{\mathsfit}{bold}{\encodingdefault}{\sfdefault}{bx}{n}

\usepackage{times}
\usepackage{iclr2024_conference}
\usepackage{times}
\usepackage{epsfig}
\usepackage{graphicx}
\usepackage{amsmath}
\usepackage{amssymb}
\usepackage{comment}
\usepackage{booktabs}
\usepackage{xcolor}
\usepackage{caption}
\usepackage{subcaption}
\usepackage{graphicx}
\usepackage{comment}
\usepackage{booktabs}
\usepackage{multirow}
\usepackage{hhline}
\usepackage{makecell}
\usepackage{footmisc}
\usepackage[export]{adjustbox}
\usepackage[colorlinks,citecolor=teal]{hyperref}
\usepackage{url}
\usepackage{wrapfig}
\usepackage[ruled]{algorithm2e}
\usepackage{mwe,tikz}\usepackage[percent]{overpic}
\usepackage{enumitem}

\newcommand\blfootnote[1]{%
  \begingroup
  \renewcommand\thefootnote{}\footnote{#1}%
  \addtocounter{footnote}{-1}%
  \endgroup
}

\begin{document}
\newenvironment{squishlist}
{   \begin{list}{$\bullet$}
    { 
    \setlength{\itemsep}{0pt}
    \setlength{\parsep}{.5em}
    \setlength{\topsep}{0pt}
    \setlength{\partopsep}{0pt}
    \setlength{\leftmargin}{1.5em} 
    \setlength{\labelwidth}{1.5em}
    \setlength{\labelsep}{0.8em} } }
      {\end{list}}
\newcommand{\xiao}[1]{\textcolor{blue}{#1}}
\newcommand{\D}{{D_{\theta}}}
\newcommand{\G}{{G_{\phi}}}
\def\eg{\emph{e.g.,}} \def\Eg{\emph{E.g.,}}          
\def\ie{\emph{i.e.,}} \def\Ie{\emph{I.e.,}}          
\def\cf{\emph{c.f.}} \def\Cf{\emph{C.f.}}          
\def\etc{\emph{etc.}}                              
\def\vs{\emph{vs.}}                                
\def\etal{\emph{et al.}}                           
\newcommand{\Hline}{\Xhline{2\arrayrulewidth}}     
\newcommand{\HLINE}{\Xhline{4\arrayrulewidth}}     
\newcommand{\dash}{-\phantom{00}}                  
\newcommand{\tstrut}{\rule{0pt}{2.0ex}}            
\newcommand{\bstrut}{\rule[-0.9ex]{0pt}{0pt}}      
\newcommand{\tcite}[1]{\scriptsize{\cite{#1}}}     
\newcommand{\tcitep}[1]{\scriptsize{\citep{#1}}}   
\title{Structural Adversarial Objectives for\\Self-Supervised Representation Learning}

\author{Xiao Zhang\\
University of Chicago\\
\texttt{zhang7@uchicago.edu}
\And
Michael Maire\\
University of Chicago\\
\texttt{mmaire@uchicago.edu}
}


\newcommand{\fix}{\marginpar{FIX}}
\newcommand{\new}{\marginpar{NEW}}

\iclrfinalcopy 

\maketitle
\blfootnote{Source code is available at \href{https://github.com/xiao7199/structural-adversarial-objectives}{https://github.com/xiao7199/structural-adversarial-objectives}}
\vspace{-2pt} 
\begin{abstract}
Within the framework of generative adversarial networks (GANs), we propose objectives that task the discriminator for self-supervised representation learning via additional structural modeling responsibilities.  In combination with an efficient smoothness regularizer imposed on the network, these objectives guide the discriminator to learn to extract informative representations, while maintaining a generator capable of sampling from the domain.  Specifically, our objectives encourage the discriminator to structure features at two levels of granularity: aligning distribution characteristics, such as mean and variance, at coarse scales, and grouping features into local clusters at finer scales.  Operating as a feature learner within the GAN framework frees our self-supervised system from the reliance on hand-crafted data augmentation schemes that are prevalent across contrastive representation learning methods.  Across CIFAR-10/100 and an ImageNet subset, experiments demonstrate that equipping GANs with our self-supervised objectives suffices to produce discriminators which, evaluated in terms of representation learning, compete with networks trained by contrastive learning approaches. 
\end{abstract}

\section{Introduction}
\label{sec:intro}

Unsupervised feature learning algorithms aim to directly learn representations from data without reliance on
annotations, and have become crucial to efforts to scale vision and language models to handle real-world
complexity.  Many state-of-the-art approaches adopt a contrastive self-supervised framework, wherein a deep neural
network is tasked with mapping augmented views of a single example to nearby positions in a high-dimension embedding
space, while separating embeddings of different examples~\citep{wu2018unsupervised,he2020momentum,chen2020simple,
chen2021exploring,grill2020bootstrap,zbontar2021barlow}.  Though requiring no annotation, and hence unaffected by
assumptions baked into any labeling procedure, the invariances learned by these models are still influenced by
human-designed heuristic procedures for creating augmented views.

The recent prominence of contrastive approaches was both preceded by and continues alongside a focus on engineering
domain-relevant proxy tasks for self-supervised learning.  For computer vision, examples include learning geometric
layout~\citep{doersch2015unsupervised}, colorization~\citep{zhang2016colorful,LMS:CVPR:2017}, and inpainting~\citep{
pathak2016context,he2022masked}.  Basing task design on domain knowledge may prove effective in increasing learning
efficiency, but strays further from an alternative goal of developing truly general and widely applicable
unsupervised learning techniques.

Another family of approaches, coupling data generation with representation learning, may provide a path toward such
generality while also escaping dependence upon the hand-crafted elements guiding data augmentation or proxy task
design.  Generative adversarial networks (GANs)~\citep{GANs,goodfellow2020generative} and variational autoencoders
(VAEs)~\citep{kingma2013auto} are prime examples within this family.  Considering GANs, one might expect the
discriminator to act as an unsupervised representation learner, driven by the need to model the real data distribution
in order to score the generator's output.  Indeed, prior work finds that some degree of representation learning
occurs within discriminators in a standard GAN framework~\citep{radford2015unsupervised}.  Yet, to improve generator
output quality, limiting the capacity of the discriminator appears advantageous~\citep{arjovsky2017wasserstein} --
a choice potentially in conflict with representation learning.  Augmenting the standard GAN framework to separate
encoding and discrimination responsibility into different components~\citep{donahue2017adversarial,
dumoulin2017adversarially}, along with scaling to larger models~\citep{donahue2019large}, are promising paths
forward.

\begin{figure*}[t]
   \centering
   \begin{subfigure}[t]{0.49\textwidth}
      \centering
      \captionsetup{justification=centering}
      \includegraphics[width=\textwidth]{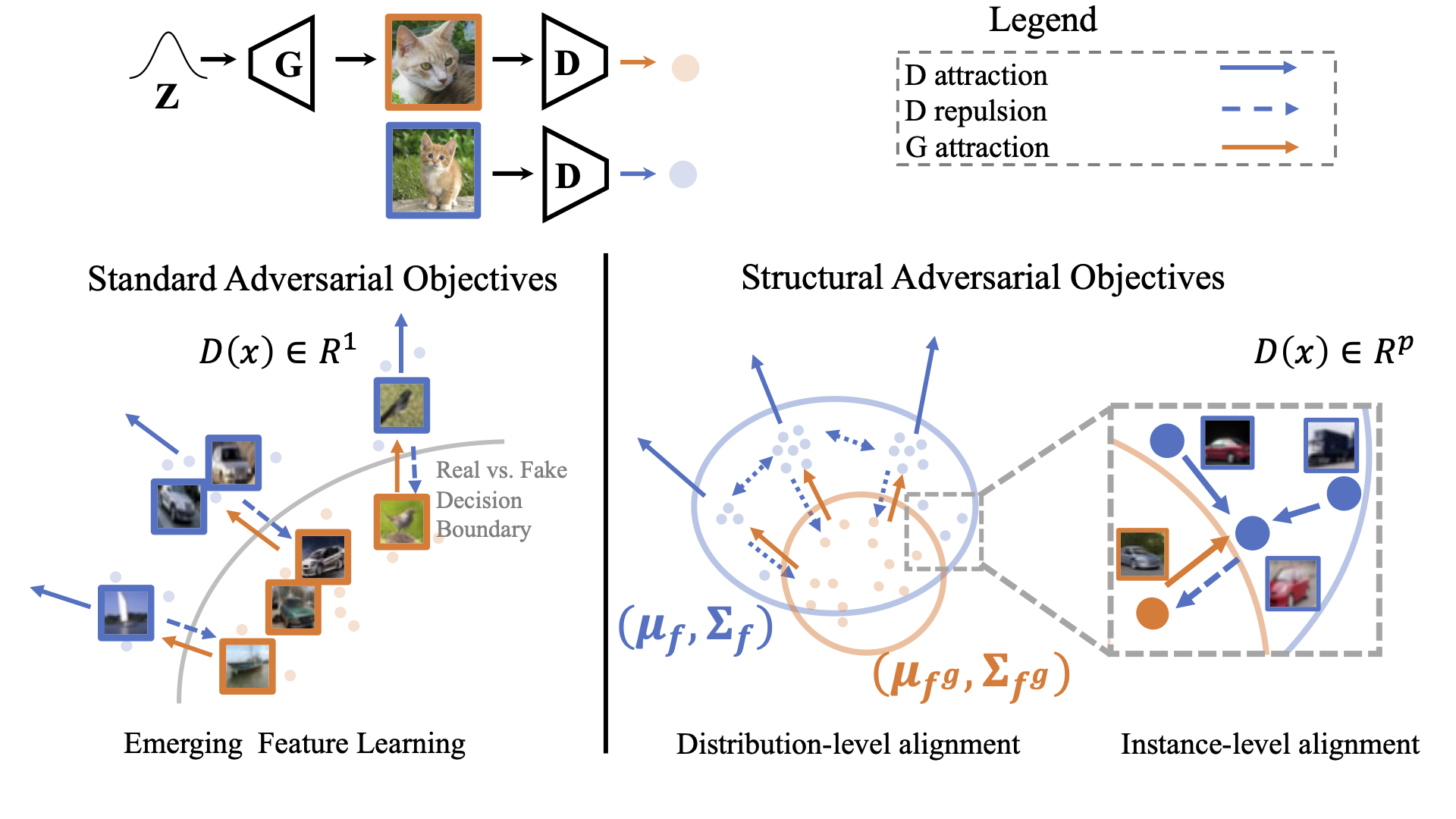}
      \vspace{-10pt}
      \caption{Standard \vs~proposed structural adversarial objectives for feature learning}
      \label{fig:overview_eval}
   \end{subfigure}
   \hfill
   \begin{subfigure}[t]{0.49\textwidth}
      \centering
      \captionsetup{justification=centering}
      \includegraphics[width=\textwidth]{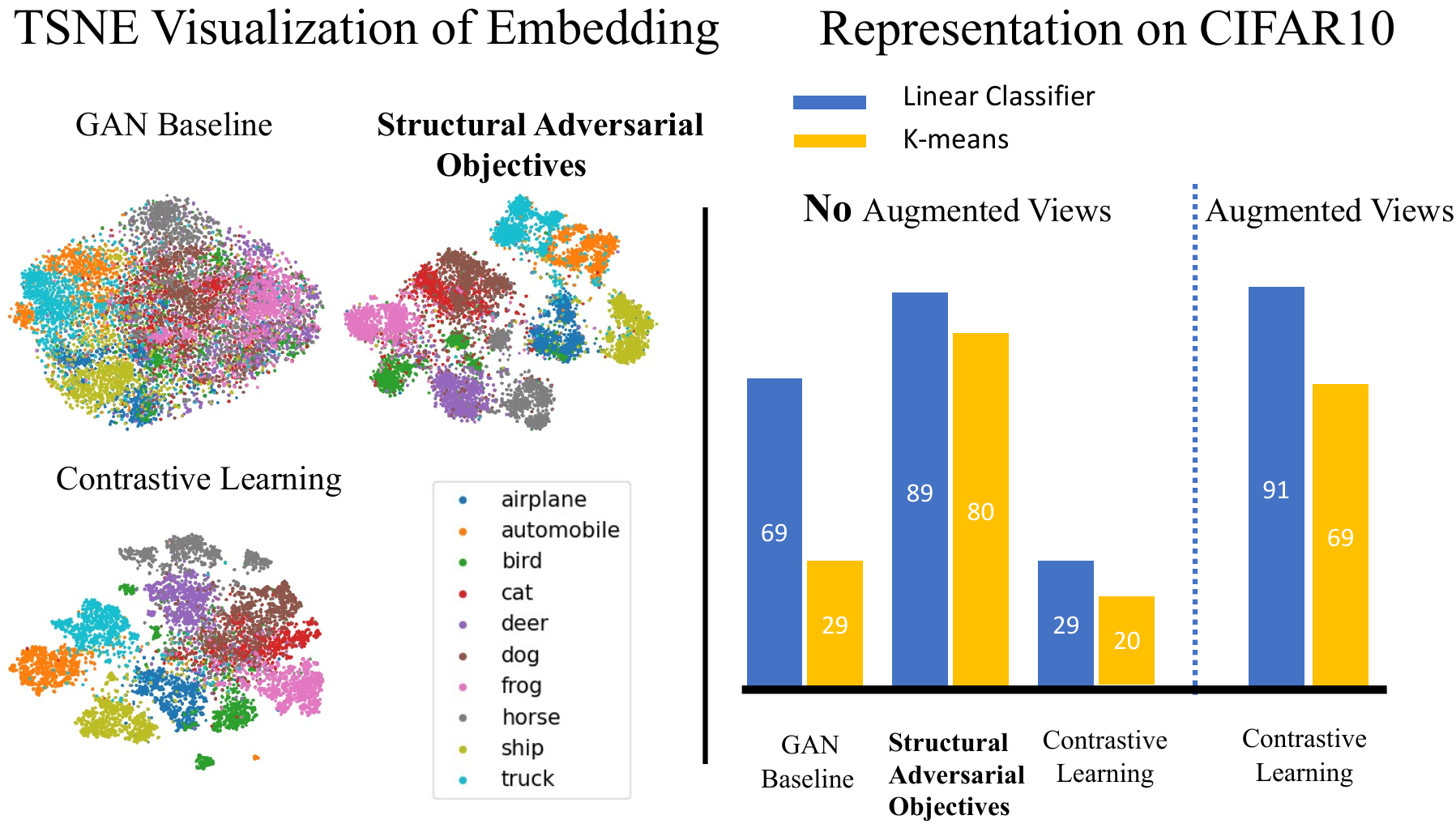}
      \vspace{-10pt}
      \caption{Visualizing and quantitatively evaluating discriminator features on CIFAR-10}
      \label{fig:overview_object}
   \end{subfigure}
   \caption{%
      \emph{(a) Structural GAN Objectives:}
      In a standard GAN, the discriminator produces a scalar score to discern real and fake samples.  As the generator
      improves, representations produced by the discriminator will update structurally similar data in a similar
      direction, displayed as solid blue arrows.  Our structural adversarial objectives enhance such learning capability
      by optimizing the feature vectors produced by the discriminator.  We achieve this by manipulating mean and
      variance at a coarser scale and implementing instance-level grouping at a finer scale, allowing the discriminator
      to explicitly learn semantic representations, in addition to distinguishing between real and fake.
      \emph{(b) Discriminator as Semantic Representation Learner:}
      Trained with our new objectives, the discriminator's learned feature embedding reveals category semantics and
      achieves performance competitive with contrastive learning methods.  Unlike self-supervised contrastive methods,
      our approach \emph{does not} depend upon learning from different views obtained via a data augmentation scheme.
   }%
   \label{fig:overview}
\end{figure*}

However, it has been unclear whether the struggle to utilize vanilla GANs as effective representation learners stems
from inherent limitations of the framework.  We provide evidence to the contrary, through an approach that
significantly improves representations learned by the discriminator, while maintaining generation quality and
operating with a standard pairing of generator and discriminator components.  To enhance GANs into effective
representation learners, our approach need only modify the training objectives within the GAN framework.  Our
contributions are as follows:
\begin{squishlist}
\item We propose adversarial objectives resembling a contrastive clustering target (Figure~\ref{fig:overview}).  These
self-supervised objectives prompt the discriminator to learn semantic representations, without depending on data
augmentation to fuel the learning process.
\item We introduce an effective regularization approach that utilizes the approximation of the spectral norm of the
Jacobian to regulate the smoothness of the discriminator.  This methodology enables the discriminator to strike a
balance between its capacity to learn features and its ability to properly guide the generator.
\item On representation learning benchmarks, our method achieves competitive performance with recent state-of-the-art
contrastive self-supervised learning approaches, even though we do not leverage information from (or even have a
concept of) an augmented view.  We demonstrate that supplementing a GAN with our proposed objectives not only enhances
the discriminator as a representation learner, but also improves the quality of samples produced by the generator.
\end{squishlist}

\section{Related Work}
\label{sec:related}

\subsection{Generative Feature Learning}
GANs~\citep{GANs,goodfellow2020generative} include two learnable modules: a generator $\G$, which produces synthetic
data given a sample $\vv$ from a prior, and a discriminator $\D$, which learns to differentiate between the true data
$\vx$ and generated samples $\G(\vv)$.  Here, $\theta, \phi$ denote the trainable parameters.  During training, $\G$
and $\D$ are alternatively updated in an adversarial fashion, which can be formulated as a minimax problem:
\begin{eqnarray}
   \min_{\G}\max_{\D}
      \mathbb{E}_{\vx\sim p(\vx)}[\log \D(\vx)] - \mathbb{E}_{\hat{\vx}\sim \G(\vv)}[1 - \log \G(\hat{\vx})].
   \label{eqn:gan_vanilla}
\end{eqnarray}
Much research on GANs has focused on improving the quality of generated data, yielding significant advances~\citep{
karras2017progressive,karras2019style,karras2020analyzing,karras2021alias,sauer2022stylegan,dai2022closedloop}.  Other
efforts have focused on evolving capabilities, including conditional and controllable generation, \eg~text-guided
\citep{zhang2021cross,hinz2020semantic} or segmentation-guided \citep{zhu2017unpaired,chen2017photographic} generation.
In comparison, adopting GANs for unsupervised feature learning has been more scarcely explored.  In this area, an
adversarial approach dependent upon an additional encoder component~\citep{donahue2017adversarial,
dumoulin2017adversarially,donahue2019large,jahanian2021generative} appears most successful to date.  Here, the encoder
is tasked to invert the generator with a discriminator acting on (data, latent) pairs and representation learning is
the responsibility of the encoder, rather than the discriminator.

Besides GANs, other generative models also demonstrate feature learning capability.  Recent efforts~\citep{
zhang2022improving,ma2021decoupling} discard the low-level structures in VAE and Flow models to improve
learned representations.  \cite{du2021unsupervised} show that an unsupervised energy model can learn semantic
structures, \eg~segmentation and viewpoint, from images.  \cite{preechakul2022diffusion} attach an encoder to a
diffusion model and show that it learns high-level feature representations.  We adopt an orthogonal approach that, by
imposing structural adversarial objectives in GAN training, tasks the discriminator to learn richer data
representations.

\subsection{Contrastive Self-Supervised Learning}
Self-supervised learning with a contrastive approach has shown enhanced feature learning capability and has evolved to
nearly match the performance of its supervised counterparts.  From initial impactful results in vision and language
\citep{wu2018unsupervised,he2020momentum,chen2020simple,radford21a}, this technique has recently been employed across a
variety of domains~\citep{jiang2022embed,krishnan2022self,guldenring2021self}.  A popular strategy involves using a
Siamese architecture to optimize the InfoNCE objective, which aims to maximize the feature similarity across augmented
views, while repulsing from all other instances to maintain feature uniformity \citep{wu2018unsupervised,
he2020momentum,chen2020simple,oord2018representation}.  Another strategy simplifies this pipeline by dropping the
negative terms and leveraging specific architectural designs to prevent collapsed solutions \citep{chen2021exploring,
grill2020bootstrap}.  As an alternative to operating on an $l_2$ normalized embedding, other approaches~\citep{
caron2020unsupervised,caron2021emerging,wang2021self} enforce clustering consistency across views.  Inspired by masked
language modeling, \cite{he2022masked} and \cite{bao2021beit} propose variants in the image domain by tasking an
autoencoder to predict masked pixels.

Though contrastive approaches yield strong benchmark results, \cite{tian2020makes} showcase the limitations of
view-invariant assumptions and demonstrate their sensitivity to the parameters of augmentation schemes.
\cite{zhang2020self} raise a concern with applying these methods to broader unconstrained datasets, where multiple
object instances within the same image should not have mutually invariant representations.

\subsection{Stabilizing GAN Training}
Despite the ability to generate high-quality samples, successfully training GANs remains challenging due to the
adversarial optimization.  Several approaches have been proposed to stabilize training and enable scaling to larger
models.  \cite{heusel2017gans} suggest maintaining separate learning rates for the generator and discriminator, in order
to maintain local Nash equilibrium.  \cite{arjovsky2017wasserstein} and \cite{gulrajani2017improved} consider
constraining the discriminator's Lipschitz constant with gradient clipping and gradient norm penalization.  In contrast
to regularizing model-wise functionality, \cite{miyato2018spectral} implement layer-wise spectral normalization schemes
by dividing parameters with their leading singular value, which is widely adopted in recent state-of-the-art models.
\cite{wu2021gradient} and \cite{bhaskara2022gran} instead propose to build a Lipschitz-constrained function by
dividing the output with the gradient norm, and show it can preserve model capacity.  However, none of these methods
suit our case, since spectral normalization \citep{miyato2018spectral} harms model capacity, and gradient-based
regularization only works for scalar output, limiting the use of structural objectives.

\section{Method: Feature Learning with the Discriminator}
\label{sec:method}

\begin{figure*}[!t]
   \vspace{-5pt}
   \centering
   \begin{minipage}[t]{0.33\textwidth}
      \vspace{5pt}
      \begin{subfigure}{.45\linewidth}
         \centering
         \captionsetup{justification=centering}
         \includegraphics[width=\linewidth]{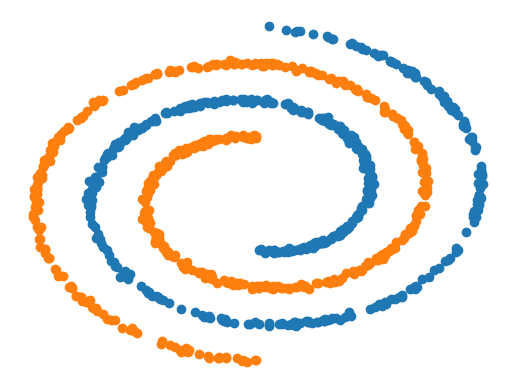}
         \caption{\small{Data\\samples}}
         \label{fig:synthetic_data_samples}
      \end{subfigure}
      \vspace{10pt}
      \begin{subfigure}{.45\linewidth}
         \centering
         \captionsetup{justification=centering}
         \includegraphics[width=\linewidth]{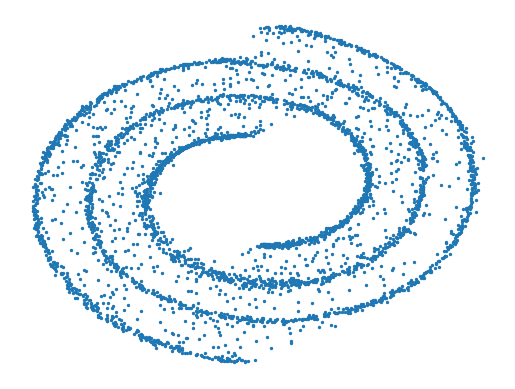}
         \caption{\small{Generated samples}}
         \label{fig:synthetic_generated_samples}
      \end{subfigure}
      \begin{subfigure}{.45\linewidth}
         \centering
         \captionsetup{justification=centering}
         \includegraphics[width=\linewidth]{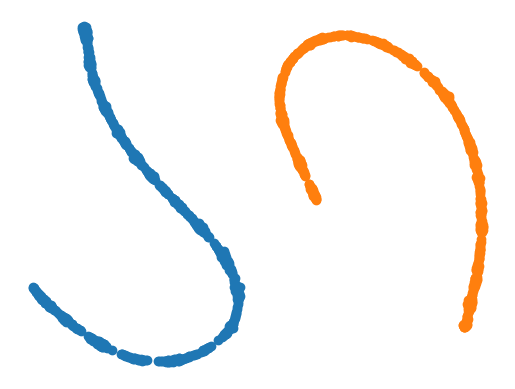}
         \caption{\small{t-SNE visualization}}
         \label{fig:synthetic_projected_embedding}
      \end{subfigure}\hspace{11pt}%
      \begin{subfigure}{.45\linewidth}
         \captionsetup{justification=centering}
         \centering
         \includegraphics[width=\linewidth]{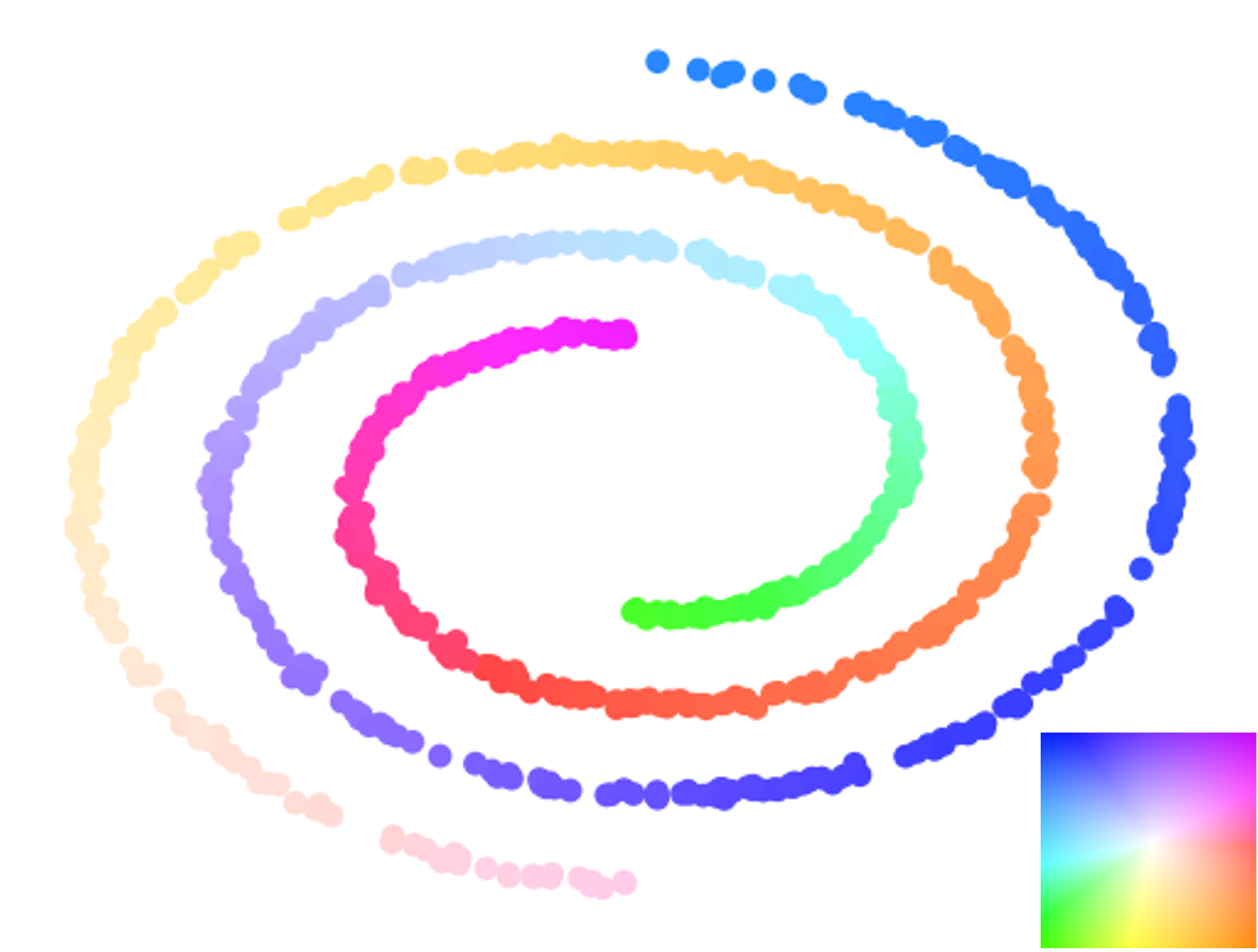}
         \caption{\small{t-SNE palette map}}
         \label{fig:synthetic_color_embedded}
      \end{subfigure}
   \end{minipage}
   \hfill
   \begin{minipage}[t]{0.325\textwidth}
      \vspace{2pt}
      \begin{subfigure}[t]{\linewidth}
         \centering
         \includegraphics[width=\linewidth]{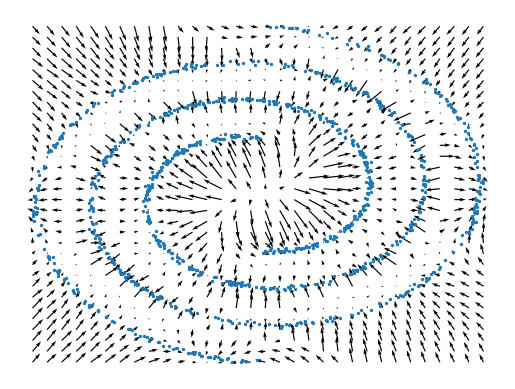}
         \begin{minipage}{0.9\linewidth}
         \caption{%
            Gradient ($\frac{\partial{\mathcal{L}}}{\partial x}$) resembles the path of optimal transport,
            suggesting $\D(x)$ can represent intrinsic structure of the data.%
         }%
         \label{fig:synthetic_data_learned_ot}
         \end{minipage}
      \end{subfigure}%
   \end{minipage}
   \hfill
   \begin{minipage}[t]{0.325\textwidth}
      \vspace{0pt}
      \begin{subfigure}[t]{\linewidth}
         \centering
         \includegraphics[width=\linewidth]{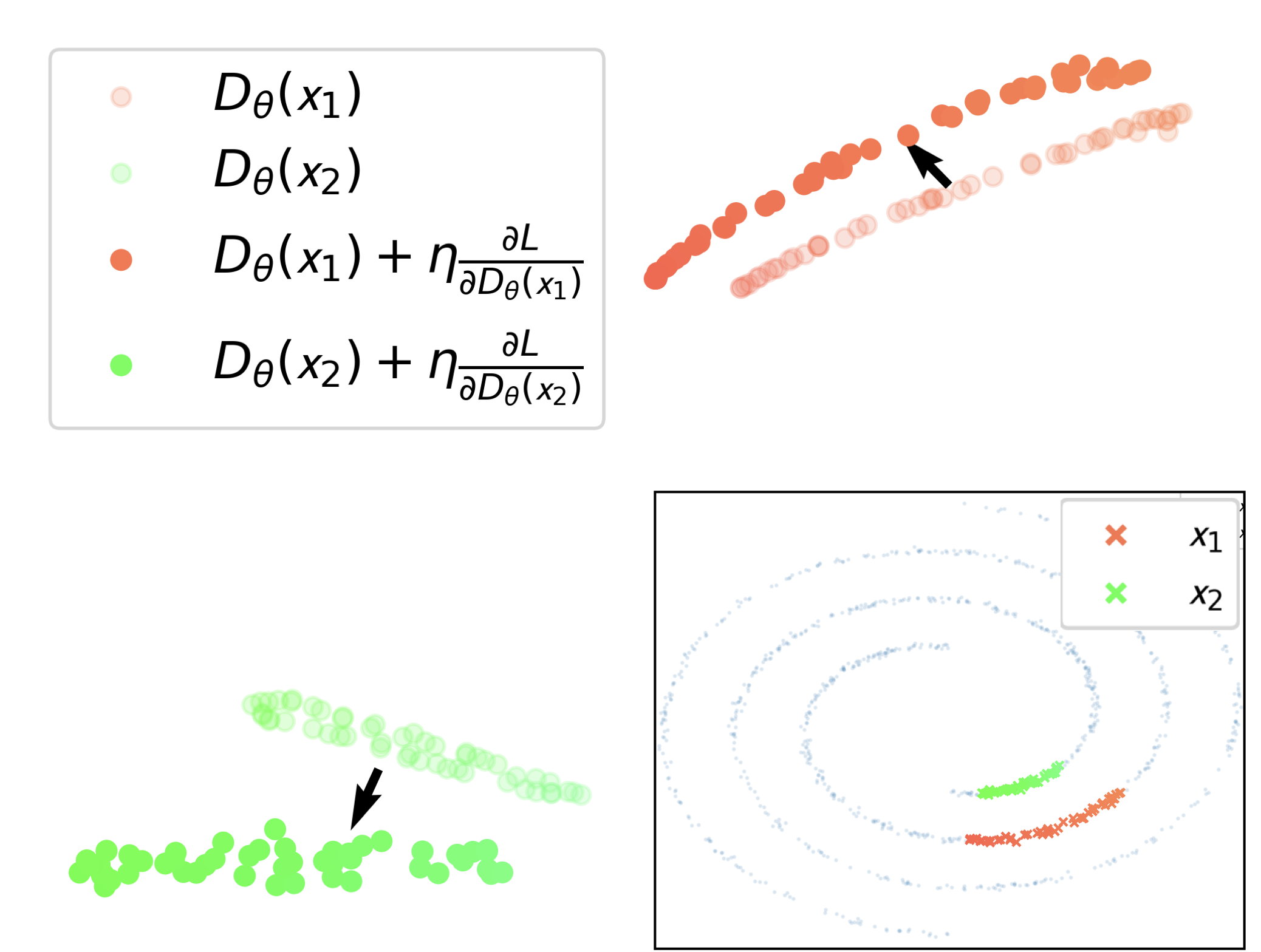}
         \caption{%
            When updated using $\mathcal{L}$, represented by arrow directions, $\D(x)$ will align with semantically
            similar data, marked by the same color, and diverge from dissimilar data, indicated by different colors.%
         }%
         \label{fig:synthetic_data_emerging}
      \end{subfigure}
   \end{minipage}
   \caption{%
      We train a GAN with our structural objectives on a synthetic \emph{double spiral} dataset.
      We show:
         \emph{(a) training data} color-coded based on ground truth assignments;
         \emph{(b) generated samples};
         \emph{(c,d) learned representations} visualized by t-SNE~\citep{van2008visualizing}, and colored according to
         ground truth categories \emph{(c)} as well as a 2D palette map \emph{(d)}.
      Additionally, we highlight \emph{(e) structural correspondence} of $\D(x)$ via
         $\frac{\partial{\mathcal{L}}}{\partial x}$, and in
      \emph{(f)}, we visualize $\D(\vx)$ using t-SNE, showcasing the emerging capability for learning semantic
      features induced by our loss $\mathcal{L}$ (Eq.~\ref{eqn:loss_overall}).%
   }%
   \label{fig:synthetic_exps}
\end{figure*}

Our goal is to task $\D$ as both a discriminator and a feature extractor that learns semantic representations of real
data.  We motivate this design from empirical observations of GAN discriminator behavior.
Figure~\ref{fig:overview_eval} conveys some intuition behind our design, while Figure~\ref{fig:synthetic_exps}
illustrates results, as well as discriminator learning dynamics when applying our method to a synthetic dataset.

As Figure~\ref{fig:synthetic_data_emerging} shows, the updating direction induced by our loss enables $\D(\vx)$ to
position example $\vx$ close to similarly structured examples while diverging away from dissimilar ones.  Such
behavior is not necessarily limited to our system; we hypothesize that it arises in broader contexts due to a
Lipschitz-regularized discriminator producing gradients that rearrange the embedding along an optimal transport path,
as shown in Figure~\ref{fig:synthetic_data_learned_ot}.  As a consequence, structurally similar samples will be
updated in a similar direction.  \cite{tanaka2019discriminator} establishes this idea in the context of Wasserstein
GANs~\citep{arjovsky2017wasserstein}.

This conjecture suggests that, in a standard GAN, the discriminator implicitly learns some, but perhaps not all,
aspects of a semantic representation.  We are therefore motivated to propose explicit objectives for the discriminator
that are both compatible with its original purpose (providing informative gradients to the generator) and that require
it to produce an embedding that captures additional semantic structure of the data distribution.

\subsection{Structural Adversarial Objectives}

Instead of producing a scalar output, we architect $\D$ to learn the mapping from the data space to the feature space,
$\D: \mathcal{X} \rightarrow \mathcal{Z}$.  We denote the output from $\D$ on real data and fake (generated) samples as
$\vz$ and $\vz^g$, respectively.  Here $\vz, \vz^g \in \mathbf{S}^{p-1}$ are normalized and live in a unit hypersphere.
We also maintain unnormalized counterparts $\tilde{\vz}$ and $\tilde{\vz}^g$ of $\vz$ and $\vz^g$;
Section~\ref{sec:smoothing} explores their utility.

Driving the formulation of our proposed objectives is the idea to require $\D$ to model the real and fake distributions
(without collapse), while $\G$ adversarially attempts to align these distributions.  As related prior work,
OT-GAN~\citep{salimans2018improving} proposes explicit optimal-transport adversarial objectives for this purpose,
but requires a large batch size (8K) to stabilize.  Instead, our objectives operate hierarchically and regularize the
learned embeddings at two levels of granularity:
\vspace{-5pt}
\begin{itemize}[leftmargin=0.3in]
\item[(1)]{%
   At a coarse level, we align the distribution statistics of the discriminator, focusing on its mean and covariance:
   $\vmu_\vz, \vmu_{\vz^g} \in \mathbb{R}^{p}$ and $\mSigma_\vz, \mSigma_{\vz^g} \in \mathbb{R}^{p\times p}$.
   Here, we simplify the optimization by assuming a diagonal structure of the covariance matrix.  This enables efficient
   alignment of the two distributions with tolerance to finer-grained differences.%
}%
\vspace{-5pt}
\item[(2)]{%
   At a finer level, we focus on reorganizing embeddings by constructing clusters using local affinity.  The
   corresponding objective tasks $\D$ with learning local geometry, further focusing the GAN on feature alignment
   between real and fake distributions.
}
\end{itemize}

\noindent
\textbf{Coarse-scale optimization by aligning distributions.}
To align the distributions in terms of mean and covariance, we can employ a distance function $d(\cdot)$ and optimize
the minimax objective:
\begin{eqnarray}
   \mathcal{L}_{\rm{Gaussian}} := \min_\G \max_\D d(\vz, \vz^g).
   \label{eqn:Gaussian}
\end{eqnarray}
One widely adopted candidate for $d(\cdot)$ is Jensen-Shannon divergence (${\rm JSD}$) due to its symmetry and stability.
For two arbitrary probability distributions $P,Q$, ${\rm JSD}$ admits the following form:
\begin{eqnarray}
   {\rm JSD}(P||Q)
    =& \frac{1}{2}(D_{\rm KL}(P || \frac{P + Q}{2}) + D_{\rm KL}(Q || \frac{P + Q}{2}))
    = \frac{1}{2}(H(\frac{P+Q}{2}) - \frac{1}{2}\left(H(P) + H(Q)\right)).
\end{eqnarray}
where $D_{\rm KL}, H$ denote Kullback-Leibler divergence and entropy, respectively.  We can compute entropy for $Q, P$
using closed-form expressions.  However, entropy for $(P + Q)/2$ is difficult to compute exactly and generally requires
Monte Carlo simulation, an infeasible computational approach in high dimensional space.  To tackle this problem, we
follow \cite{hershey2007approximating} to approximate $\frac{P + Q}{2}$ by a single Gaussian and estimate sample mean
and covariance by joint samples of $P$ and $Q$, which yields an upper bound of $H(\frac{P+Q}{2})$; the bound is
tight when $P = Q$.  Putting these together, we obtain our distance function for the coarser scale
objective\footnote{Note that though Eq.~\ref{eqn:JSD} and $\rm{MCR}$ in \cite{dai2022closedloop} are constructed
similarly, the latter is interpreted from a coding rate reduction perspective.}:
\begin{eqnarray}
   {\rm JSD}(\vz, \vz^g) \approx \log \frac{\det \mSigma_{\vz + \vz^g}}{\sqrt{\det \mSigma_\vz \det \mSigma_{\vz^g}}}.
   \label{eqn:JSD}
\end{eqnarray}
Another well-established  metric between two Gaussian distributions is Bhattacharyya distance $D_B$:
\begin{eqnarray}
   D_B(\vz, \vz^g) :=
      \frac{1}{8}(\vmu_\vz - \vmu_{\vz^g})^T\mSigma^{-1}(\vmu_\vz - \vmu_{\vz^g}) +
      \frac{1}{2}\log \frac{\det \mSigma }{\sqrt{\det \mSigma_\vz \det \mSigma_{\vz^g}}},
   \label{eqn:DB}
\end{eqnarray}
where $\mSigma = \frac{\mSigma_\vz + \mSigma_{\vz^g}}{2}$.
Though having different geometric interpretations, it is notable that $D_B$ and ${\rm JSD}$ have similar format and,
when maximizing $d(\vz, \vz^g)$ for $\vz$, both aim to uniformly repulse $z$ to prevent producing collapsed
representations.  In experiments, we observe that these two distances yield similar performance and we use ${\rm JSD}$
as our default choice for $d(\cdot)$ since it has a slightly faster convergence rate and yields better quality for
generated images.

\vspace{5pt}
\noindent
\textbf{Fine-grained optimization via clustering.}
We perform mean-shift clustering on $\vz$ by grouping nearby samples.  We simplify the clustering process by equally
averaging each neighbor sample, rather than using feature similarity to reweight their contribution.  To improve
nearest neighbor search stability, we maintain a rolling updated memory bank $\vz^m$ that stores the embedding of all
real images as a query pool and use the backbone representation $\vz^b$, rather than $\vz$, as the key to computing
feature similarity.  Denoting $\{\vz_{i,j}\}_{j=1}^k$ and $\{\vz_{i,j}^g\}_{j=1}^K$ as the returned $K$ nearest
neighbors of real images embedding for $\vz_i$ and $\vz_i^g$ respectively, our clustering objective is:
\begin{eqnarray}
   \mathcal{L}_{\rm{cluster}} :=
      \max_\D \frac{1}{NK}\sum_{i=1}^N \sum_{j=1}^K \vz_{i,j}^\top \vz_i +
      \min_\D \max_\G  \frac{1}{N}\sum_{i=1}^N \sum_{j = 1}^K {\vz_{i,j}^g}^\top \vz^g_i.
   \label{eqn:real_knn}
\end{eqnarray}
IC-GAN~\citep{casanova2021instance} implements a similar instance-wise objective.  However, they use frozen embeddings
from an off-the-shelf model rather than jointly learn an embedding, and their motivation is to improve image generation
quality rather than learn semantic features --- entirely different from our aim.

\subsection{Smoothness Regularization}
\label{sec:smoothing}

Besides reformulating adversarial targets for representation learning, we address another common issue in GAN training:
balancing the discriminator's capacity and the smoothness constraint.  Recent studies demonstrate that regularizing
$\D$'s smoothness, or its Lipschitz constant, is critical for scaling GANs to large network architectures.  Consider a
continuous function $F: \mathbb{R}^m\rightarrow\mathbb{R}^p$.  We can bound its Lipschitz constant by the spectral norm
of Jacobian $\mJ_F$:
   $$\|\mJ_F(\vx) \|_2 \leq \rm{Lip},$$
where $\|\cdot \|_2$ denotes the matrix spectral norm.  However, computing the full Jacobian matrix is highly
inefficient in standard backpropagation process since each backpropagation call can only compute a single row of the
Jacobian matrix, which is impracticable as we usually need large embedding dimension $p$.

\newlength{\commentWidth}
\setlength{\commentWidth}{2cm}
\newcommand{\atcp}[1]{\tcp*[r]{\makebox[\commentWidth]{#1\hfill}}}
\newlength{\oldintextsep}
\setlength{\oldintextsep}{\intextsep}
\setlength\intextsep{0ex}
\begin{wrapfigure}[14]{rT}{0.48\textwidth}
   \begin{minipage}{0.48\textwidth}
   \begin{algorithm}[H]
      \caption{Approximating $\|\mJ_F(\vx)\|_2$ with power iterations}
      \textbf{Input:}
         Function $F: \mathbb{R}^m\rightarrow\mathbb{R}^p$;
         Stop gradient operator $sg(\cdot)$;
         Power iteration steps S;
         Batch size $b$;
         Input data $\vx\in \mathbb{R}^{b\times m}$;\\
      Init random vector $\vu\sim \mathcal{N}(0, 1) \in \mathbb{R}^{b\times p}$\\
      \For{iter $= 1\dots S$ }{
         $\vv = \vu \mJ_F(\vx)/ \|\vu \mJ_F(\vx)\|_2 \,\,\,\,\,\,\,\,//\texttt{VJP}$ 
         \\
         $\vu = \mJ_F(\vx)\vv/ \|\mJ_F(\vx)\vv\|_2\,\,\,\,\,\,\,\,\,//\texttt{JVP}$ 
      }
      \textbf{Return:} $\|\mJ_F(\vx)\|_2 \approx sg(\vu) \mJ_F(\vx) sg(\vv)$
      \label{alg}
   \end{algorithm}
\end{minipage}
\end{wrapfigure}

Therefore, we propose to efficiently approximate $\|\cdot \|_2$ using power-iterations.  Leveraging the fact that
power-iteration is a matrix-free method, we do not need to explicitly compute the Jacobian matrix.  Instead, we only
need to access the matrix by evaluating the matrix-vector product, which can be efficiently computed by batch-wise VJP
and JVP (Jacobian-Vector-Product) subroutine.  Algorithm~\ref{alg} presents the details, where only $(2S+1)$
backpropagation calls are required to approximate $\|\mJ_\D(\vx)\|_2$.  In experiments, we find that $S=1$ suffices for
a ResNet-18 model.

We observe that maintaining $\|\tilde{\vz}\|$ at regular level benefits training stability.  To this end, we use hinge
loss to regularize the embedding norm and empirically observe it performs better than removing the hinge.  Therefore,
our smoothness regularization is:
\begin{eqnarray}
   \min_\D\mathbb{E}_{\vx}\|\mJ_\D(\vx) - {\rm Lip}\|_2 +
   \lambda_{h}\mathbb{E}_{\boldsymbol{\tilde{z}}}\|\max(\| \boldsymbol{\tilde{z}} \| - 1, 0)\|_2
\end{eqnarray}
where $\lambda_h$ denotes the ratio for hinge regularization, and \rm{Lip} denotes the Lipschitz target of $\D$, which
is set to 1 by default.  Unlike a layer-wise normalization scheme, \eg~Spectral Norm~\citep{miyato2018spectral}, where
demanding local regularization hurts the model's capacity, our proposed regularization scheme allows the network to
simultaneously fit multiple objectives, \ie~representation learning and smoothness regularization.  The model does not
have to sacrifice capacity for smoothness.  Another benefit of our method is that our proposed term can work with the
normalization layer.  Spectral Norm cannot, because of the data-dependent scaling term in its normalization layer.

\vspace{2pt}
\noindent
\textbf{Overall training objective.} We define our final objective as:
\begin{eqnarray}
   \mathcal{L} :=
      \mathcal{L}_{\rm{Gaussian}} + \lambda_c\mathcal{L}_{\rm{cluster}}  +  \lambda_s \mathcal{L}_{\rm{reg}},
   \label{eqn:loss_overall}
\end{eqnarray}
where $\lambda_c, \lambda_s$ control the relative loss weights.

\section{Experimental Settings}
\label{sec:exp}

We train our model for 1000 epochs on CIFAR-10/100 and 500 epochs on ImageNet-10.  We use the AdamW optimizer~\citep{
loshchilov2017decoupled} with a constant learning rate of 2e-4 for both generator and discriminator.  We additionally
add 0.1 weight decay to the discriminator.  We use batch size 500 on CIFAR-10/100 and 320 on ImageNet-10.  We run a
small-scale parameter tuning experiment for hyperparameters and find that setting
   $\lambda_h = 4, \lambda_{c} = 3, \lambda_{s} = 5$
yields the best result.  For simplicity, we run a single discriminator update before optimizing the generator,
\ie~$n_{dis} = 1$.

As a widely adopted GAN training trick, we maintain a momentum-updated discriminator and generator for evaluation
purposes and find they produce stable data representations and better quality images.  We also try producing $\vz^b$
from momentum models for nearest neighbor searching, which slightly improves performance in all benchmarks.  We set
the memory bank size $|\vz^m|$ = 10240, which is smaller than all datasets, preventing the model from accidentally
picking features from augmented versions of the input image.  Appendix~\ref{sec:appendix_details} provides more
model configuration details.

\section{Results And Discussion}
\label{sec:results}

\subsection{Synthetic Data}
\label{sec:results_synthetic}

For illustrative purposes, we first train a GAN using our structural objectives on the synthetic double spirals
dataset~\citep{li2022neural}.  Here, we implement discriminator and generator as multi-layer perceptrons and keep all
other configuration, \eg~normalization layers, activation functions, objectives, and learning rate, consistent with
our settings for experiments on real images.

Figure~\ref{fig:synthetic_data_samples} demonstrates that the generated samples capture all data modes, with few
outlier samples between spirals.  Besides generation capability, we also visually inspect the discriminator's
learned representations using t-SNE~\citep{van2008visualizing}.  Figure~\ref{fig:synthetic_projected_embedding} shows
embeddings of the two categories are substantially separated.  Figure~\ref{fig:synthetic_color_embedded} colors each
data point by projecting its learned representation into a 2d palette map.  From this plot, we see that the learned
embedding preserves semantic structure within and across groups.  Figure~\ref{fig:synthetic_data_learned_ot} shows the
gradient of embedding distance approximates the optimal paths between uniform grids and data samples, indicating $\D$
learns intrinsic data structure.  Figure~\ref{fig:synthetic_data_emerging} demonstrates the capability of our
structural objectives to learn semantic features: when the embedding is updated via $\mathcal{L}$, data that are
semantically similar are updated to align in the same directions, whereas data from different clusters diverge.

\subsection{Representation Learning on Real Images}
\label{sec:results_images}

We task the backbone of the discriminator to produce a vector as a data representation and then evaluate its
performance on the task of image classification.  We compare the results with state-of-the-art contrastive learning
approaches under two widely adopted evaluation metrics:
\begin{squishlist}
   \item \textit{Linear Support Vector Machine (SVM)}:
   We optimize a Linear SVM on top of training feature and report the accuracy on the validation set.
   \item \textit{K-Means clustering}:
   We run spherical K-means clustering on the validation set, with K equaling the number of ground-truth categories.
   We then obtain a prediction on the validation set by solving the optimal assignment problem between the partition
   produced by clustering and the ground-truth categories.  To reduce the randomness in clustering, we repeat this
   process 20 times and report average performance.
\end{squishlist}

\begin{table}[]
   \begin{center}
   \setlength{\tabcolsep}{5.5pt}
   \begin{tabular}{@{}l|c|c|c|c|c|c|c}
      \multirow{2}{*}{Method} & \multirow{2}{*}{Parameters (M)} & \multicolumn{2}{c|}{CIFAR-10} & \multicolumn{2}{c|}{CIFAR-100} & \multicolumn{2}{c}{ImageNet-10}\\
          & & SVM & K-M & SVM & K-M & SVM & K-M\\
      \HLINE
      \vspace{-10pt}&&&&&&\\
      Supervised &  11.5 & 95.1 & 95.1  & 75.9 & 73.6  & 96.4 & 96.3 \\
      Random &  11.5 & 42.9 & 22.0 & 18.3 & 8.9 & 48.2 & 28.3  \\
      \hline
          {\text DINO~\citep{caron2021emerging}} & 11.5 & 89.7 & 63.9  & 65.6 & 36.7 & 87.8 & 68.0\\
          {\text NNCLR~\citep{dwibedi2021little}} &  11.5 & 91.7 & 69.3 & 69.7 & 40.4 & \textbf{91.4} & 66.8\\
          {\text SimCLR~\citep{chen2020simple}} &  11.5 &90.6 & 75.3 & 65.6 & 41.3 & 89.0 & 65.7\\
          {\text BYOL~\citep{grill2020bootstrap}} &11.5& \textbf{93.1} & 75.0 & \textbf{70.6} & \textbf{42.8} & 90.4 & 67.3\\
          {\text SWAV~\citep{caron2020unsupervised}} &11.5&  89.1 & 64.5  & 65.0 & 35.2 & 90.0 & 61.9\\
      \hline
      MAE~\citep{he2022masked} & 20.4 &82.3 & 37.0 & 57.1 & 17.9 & 88.4 & 45.8 \\
      DDPM~\citep{ho2020denoising}&41.8& 91.1 &  78.0 & 62.5 & 36.3 & - & - \\
      Ours & 11.5&89.8 & \textbf{80.1} & 63.3 & 38.2 & 91.2 & \textbf{75.4} \\
   \end{tabular}
   \end{center}
\vspace{-10pt}
\caption{%
\emph{Representation Learning Performance.}
We evaluate our trained discriminator by benchmarking its learned representation using linear SVM and K-Means
clustering (K-M), reporting average accuracy over 20 runs.  Our method, which does not leverage any augmented views,
achieves competitive performance with self-supervised approaches across multiple datasets.  Compared to denoising
autoencoders (shown in the penultimate and antepenultimate rows), our method excels in learning more effective
representations while utilizing fewer parameters.%
}%
\label{tab:main_table}
\end{table}

Table~\ref{tab:main_table} reports results and provides comparison with current state-of-the-art methods.  For
datasets with fewer categories, \ie~CIFAR-10 and ImageNet-10, our method significantly outperforms all contrastive
learning approaches on the K-means clustering metric.  On CIFAR-10, we achieve 80.1\% test accuracy, surpassing the
best-competing method, {\text SimCLR}, which achieves 75.3\% test accuracy.  On ImageNet-10, our method reaches 75.4\%
test accuracy surpasses the best-competed method, {\text DINO}, with 68.0\% test accuracy.  When evaluating learned
representations using linear SVM, our method reaches 89.8\% test accuracy, which exceeds {\text SWAV} and {\text DINO},
with 89.1\% and 89.7\% test accuracy respectively, but falls slightly behind BYOL (93.1\%) and {\text NNCLR} (91.7\% test
accuracy).  On ImageNet10, our method's 91.2\% accuracy approaches that of the best method ({\text NNCLR} with 91.4\%)
and exceeds the rest.

CIFAR-100 contains fewer training samples per category and operationalizing instance-wise discriminating
objectives is thus favorable over clustering objectives or smoothness regularization.  Under such case, our method
remains competitive on the linear SVM metric, achieving 63.3\% test accuracy, which is very close to {\text DINO},
{\text SimCLR}, and {\text SWAV}, which each have around 65\% test accuracy.  Using K-Means clustering, our method
reaches 38.2\% test accuracy, outperforming clustering-based contrastive approaches {\text SWAV} (35.2\%) and
{\text DINO} (36.7\%).

\begin{table*}[]
   \begin{center}
   \setlength{\tabcolsep}{4.5pt}
   \begin{tabular}{@{}l|l|cccc|cccc@{}}
   \multirow{2}{*}{Data Aug} & \multirow{2}{*}{Method} & \multicolumn{4}{c|}{CIFAR-10} & \multicolumn{4}{c}{CIFAR-100}  \\
   && KMeans & SVM & LP & 1/5 KNN & KMeans & SVM & LP & 1/5 KNN\\
   \HLINE
   \vspace{-10pt}&&&&&&&&&\\
   None & SimCLR & 14.2 & 21.8 & 21.8 & 14.7/15.1 & 2.1 & 4.5 & 5.4 & 2.7/2.5 \\
   None & Ours & \textbf{76.5} & \textbf{84.5} & \textbf{83.2} & \textbf{79.7}/\textbf{82.2}  & \textbf{22.5} & \textbf{52.2} & \textbf{51.6} & \textbf{37.5} / \textbf{37.4}\\
   \hline
   $F$ & SimCLR & 17.5 & 32.8 & 32.1 & 23.3 / 25.5 & 4.1 & 10.3 & 10.9 & 6.0/5.5\\
   $F$ & Ours & \textbf{76.2} & \textbf{85.7} & \textbf{85.8} & \textbf{80.9} / \textbf{84.5} & \textbf{30.8} & \textbf{52.0} & \textbf{56.5} & \textbf{41.0} / \textbf{42.9}\\
   \hline
   $F + C$ & SimCLR & 27.8 & 72.7 & 72.2 & 64.9 / 67.2 & 12.2 & 35.1 & 34.3 & 30.8 / 29.1\\
   $F + C$ & Ours & \textbf{80.0} & \textbf{89.3} & \textbf{88.4} & \textbf{87.7} / \textbf{89.2} &   \textbf{37.4} & \textbf{63.2} & \textbf{62.0} & \textbf{55.0} / \textbf{56.1}\\
   \hline
   $F + C + J$ & SimCLR & 78.0 & 90.7 & 90.2 & 88.1 / 89.5 & 41.7 & 65.2 & 65.2 & 59.2 / 61.4\\
   \end{tabular}
   \end{center}
   \vspace{-10pt}
\caption{%
\emph{Data Augmentation Dependence.}
We compare with SimCLR~\citep{chen2020simple} on sensitivity to various data augmentation schemes.  In our system,
data augmentation is solely employed to enlarge the training dataset; it is not used for achieving view-consistency
objectives.  $F, C, J$ denote random horizontal flipping, random image cropping, and color jittering, respectively;
None means no augmentation is applied during training.  For each augmentation scheme, our method outperforms SimCLR
across all evaluation metrics (here, LP denotes linear probing).  Moreover, we are able to operate even without data
augmentation -- a regime in which SimCLR fails.%
}%
\label{tab:ablate_data_aug}
\end{table*}

Our quantitative comparison is also qualitatively confirmed by visualizing embeddings using t-SNE.  {\text BYOL}, as
shown in Figure~\ref{fig:overview}, produces isolated and smaller-sized clusters that maintain sufficient space to
discern categories under linear transformation.  However, those clusters lack sufficient global organization, which is
a quality evaluated by the K-Means clustering metric.

\begin{wraptable}{r}{5.8cm}
   \centering
   \setlength{\tabcolsep}{8pt}
   \begin{tabular}{l|c|c}
      Method & NMI & Purity\\
      \HLINE
      \vspace{-10pt}&&\\
      Self-cond GAN & 33.26 & 11.73 \\
      \hline
      Ours & \textbf{72.77} & \textbf{81.52} \\
   \end{tabular}
\vspace{-3pt}
\caption{%
\emph{Comparison to Self-conditioned GAN~\citep{liu2020diverse} on CIFAR-10.}
On normalized mutual information (NMI) and purity metrics, our method outperforms
self-conditioned GAN~\citep{liu2020diverse}, a generative model which clusters discriminator
features iteratively in a self-discovering fashion.%
}%
\label{tab:compare_table_2}
\end{wraptable}

In contrast, our approach, as also shown in Figure~\ref{fig:overview}, produces smoother embeddings, which are nearly
aligned with the ground-truth partition, and consequently yields good K-Means clustering performance.  When compared to
denoising autoencoders such as DDPM~\citep{ho2020denoising} and MAE~\citep{he2022masked}, our model demonstrates
superior efficiency by utilizing fewer parameters (11.5M) compared to DDPM (41.8M) and MAE (20.4M).  Additionally, our
model excels in learning better representations across all evaluated metrics, with the sole exception being a
comparison to DDPM on CIFAR-10 (our 89.7\% accuracy using SVM \vs~DDPM's 91.1\%).

\begin{table}[!t]
   \centering
   \setlength{\tabcolsep}{7.75pt}
   \begin{tabular}{@{}c|c|c|c|c|c|c|c}
      \multirow{2}{*}{Method / Loss} &
        \multirow{2}{*}{D regularizer}  &
         \multicolumn{2}{c|}{Parameters (M)} &
         \multirow{2}{*}{IS $\uparrow$} &
         \multirow{2}{*}{FID $\downarrow$} &
         \multirow{2}{*}{K-Means} &
         \multirow{2}{*}{SVM}\\
          & & $D$& $G$&&&&\\
      \HLINE
      \vspace{-10pt}&&&&\\
      {\text StyleGAN2-ADA}  & Grad Penalty &20.7&19.9& \textbf{9.82} & \textbf{3.60} & 28.96 & 76.50 \\
      \hline
      {\text BigGAN}& Spectral Norm & 4.2 &4.3&8.22 & 17.50 & 29.69 & 69.31\\
      \hline
      {\text Hinge Loss} &$\mathcal{L}_{\rm{reg}}$& 11.5&4.9&8.13  & 18.54  & 36.41 & 77.19 \\
      \hline
      Eq.~\ref{eqn:Gaussian} only, $D_B$ &  $\mathcal{L}_{\rm{reg}}$ & 11.5&4.9 & 8.39  &  17.83 & 70.76 & 87.9\\
      \hline
      Eq.~\ref{eqn:Gaussian} only, $\rm{JSD}$ & $\mathcal{L}_{\rm{reg}}$& 11.5&4.9& 8.55 & 16.97 & \textbf{80.55} & 88.32\\
      \hline
     Full Objectives &Spectral Norm &11.5&4.9&  7.23 & 26.41 & 55.38 & 83.9\\
      \hline
      Full Objectives &$\mathcal{L}_{\rm{reg}}$&11.5&4.9 &  8.73  & 13.63 & 80.11 & \textbf{89.76}\\
   \end{tabular}
\vspace{-5pt}
\caption{%
\emph{Ablation over Loss Function Components on CIFAR-10.}
We compare StyleGAN2-ADA~\citep{karras2020training}, BiGAN~\citep{brock2018large} and a GAN baseline using the standard
hinge loss to models using ablated variants of our structural objectives.  Dicriminators trained using our objectives
significantly outperform these baselines (K-Means, SVM metrics), while our corresponding generators also benefit
(IS, FID).  Including our finer scale clustering objective (last row) improves both representation and image quality
over ablated variants using only our coarse scale objective (rows 4 \& 5).  The benefit observed when using
$\mathcal{L}_{\text{reg}}$ over Spectral norm (final to penultimate row) indicates that preserving model capacity
is crucial for effective feature learning.%
}%
\label{tab:img_quality_table}
\end{table}

\subsection{Ablation Experiments}
\label{sec:results_ablation}

\noindent
\textbf{Sensitivity to data augmentation.}
Though we adopt some minimal data augmentation in our experiments, our approach is far less sensitive to data
augmentation.  However, contrastive self-supervised learning approaches, including SimCLR~\citep{chen2020simple},
require a carefully calibrated augmentation scheme to achieve good performance.  Table~\ref{tab:ablate_data_aug}
highlights this discrepancy.  Our method demonstrates a clear advantage over SimCLR across all augmentation
regimes, and, unlike SimCLR, can still learn useful features when no augmentation applied.

\noindent
\textbf{Comparison to other generative feature learners.}
GenRep~\citep{jahanian2021generative} generates images pairs by sampling adjacent features in the latent space of
BigBiGAN~\citep{donahue2019large} and then trains an encoder to optimize contrastive objectives.
\begin{wraptable}{r}{7.5cm}
   \centering
   \vspace{5pt}
   \begin{tabular}{l|c|c}
      Method  & Network & Linear Probing\\
      \HLINE
      \vspace{-10pt}&&\\
      GenRep(Tz only) & ResNet-50 & 55.0*\\
      \hline
      Ours &  ResNet-18 & \textbf{59.9}\\
   \end{tabular}
\vspace{-3pt}
\caption{%
\emph{ImageNet-100.}
Our method outperforms GenRep~\citep{jahanian2021generative} though we adopt a simpler network architecture and a more
direct training pipeline.  *For a fair comparison, this result is from Figure~6 of
GenRep~\citep{jahanian2021generative}, which does not use data augmentation (Tz only).}
\label{tab:compare_table}
\end{wraptable}
To compare with GenRep, we train our model on ImageNet-100, following most of the our settings for ImageNet-10, except
we extend training to 1000 epochs.  For fair comparisons, we utilize their \textit{Tz only} version, a setting where no
data augmentation is used, and show the results in Table~\ref{tab:compare_table}.  Our method outperforms GenRep,
though we adopt a simpler network architecture for feature learning.

Self-conditioned GAN~\citep{liu2020diverse} clusters the discriminator's features iteratively in a self-discovering
fashion; cluster information is fed into the GAN pipeline as conditional input.  Though this method produces
clustering during training, its objective differs entirely from ours: their motivation is to improve the diversity
of image generation, rather than learn representations.  Table~\ref{tab:compare_table_2} shows that our method
outperforms it.

\textbf{Ablation of system variants.}
Table~\ref{tab:img_quality_table} provides a quantitative comparison of both generator and discriminator performance
across baselines as well as ablated and full variants of our system.  Our proposed objectives significantly improve
generation and representation quality over the hinge loss baseline.  We witness further enhancement in image quality
when using our extra instance/clustering-wise objective.  Performance drops by replacing $\mathcal{L}_{\text{reg}}$
with spectral norm, indicating the effectiveness of our suggested regularization scheme in preserving model capacity.
As an additional advantage over spectral norm, we observed better training stability when using our regularization
scheme.  Note that while StyleGAN2-ADA achieves state-of-the-art generation quality, it both requires adopting a
larger network to do so, and still performs worse at feature learning than our system.
Appendix~\ref{sec:appendix_qualitative} provides a qualitative comparison with examples of generated images.

\section{Conclusion}
Our structural adversarial objectives augment the GAN framework for self-supervised representation learning, shaping
the discriminator's output at two levels of granularity: aligning features via mean and variance at coarser scale and
grouping features to form local clusters at finer scale.  Benchmarks across multiple datasets show that training a GAN
with these novel objectives suffices to produce data representations competitive with the state-of-the-art
self-supervised learning approaches, while also improving the quality of generated images.

{\small
\bibliographystyle{iclr2024_conference}
\bibliography{iclr2024_conference.bib}
}

\newpage
\appendix

\section{Appendix}
\label{sec:appendix}

\subsection{Details of Dataset and Model}
\label{sec:appendix_details}

\textbf{Datasets.}
We focus on three benchmark datasets: CIFAR-10, CIFAR-100 \citep{krizhevsky2009learning} and ImageNet-10.

\textit{ImageNet-10}:
We follow \cite{chang2017deep} to select 10 categories from the ImageNet dataset~\citep{5206848}, resulting
in 13,000 training images and 500 validation images.  During training, we only perform spatial augmentation, including
random spatial cropping and horizontal flipping, followed by resizing images to 128x128 resolution to match the
generated images.  During testing, we resize the images to align the smaller edge to 144 pixels, followed by central
cropping to produce a 128x128 output.

\textit{CIFAR-10/100}:
During training, we apply the same augmentation strategy as in ImageNet-10 but produce 32x32 images.  During testing,
we do not perform cropping.

For compared methods, we keep their default augmentation strategy.  On ImageNet-10, we resized their augmented images
to 128x128.  For all methods, we learn in an unsupervised manner on the training split and evaluate on the validation
split.

In CIFAR-10/100 experiments, we use default configurations from SOLO-Learns~\citep{JMLR:v23:21-1155}, an open source
library providing heavily tuned configurations for multiple state-of-the-art self-supervised methods.   In ImageNet-10
experiments, we train competing approaches using the suggested hyperparameters for ImageNet-100, but extend the total
epochs to 1000 for sufficient convergence.  For fair comparison, we run these methods with our modified backbone and
resize input images to 128x128.

\vspace{5pt}
\noindent
\textbf{Model details: discriminator.}
We construct our discriminator using ResNet-18~\citep{he2016deep} and perform several modifications to make it
cooperate reasonably with the generator.  Inspired by the discriminator configuration in BigGAN~\citep{brock2018large},
we perform spatial reduction only within the residual block and replace all stride two convolution layers with average
pooling followed by stride one convolution.  We remove the first max-pooling layer and switch the first convolution
layer to a 3x3 kernel with a 1x1 stride to keep the resolution unchanged before the residual block.

To maintain a substantial downsample rate in ImageNet-10 images, we duplicate the first residual block and enable a
spatial reduction in all blocks to reach a 32x downsampling.  On CIFAR-10/100, we preserve the default setting for
residual blocks.  As our proposed smoothness term regularizes each sample, we replace all BatchNorm
layers~\citep{ioffe2015batch} with GroupNorm~\citep{wu2018group}, specifying 16 channels as a single group; this
prevents batch-wise interaction.  We also remove the first normalization layer in each block, as doing so produces
better results.  We replace ReLU with ELU~\citep{clevert2015fast} activations for broader non-linear support on
negative values.

\vspace{5pt}
\noindent
\textbf{Model details: generator.}
We adapt the generator configuration from BigGAN-deep\citep{brock2018large}.  Specifically, we take their model for
32x32 images on CIFAR, and additionally increase the base channels to 128 to prevent image generation from being the
system bottleneck.  For ImageNet-10, we replicate their settings for 128x128 images.

\subsection{Compared Self-Supervised Learning Methods}
\label{sec:appendix_ssl}

We evaluate the representations produced by our method in comparison to those produced by the following
state-of-the-art self-supervised learning methods:

\vspace{5pt}
\begin{squishlist}
\item {\text SimCLR}~\citep{chen2020simple} optimizes the InfoNCE loss, maximizing feature similarity across views
while repulsing all the images.
\item {\text NNCLR}~\citep{dwibedi2021little} samples nearest neighbors from the data set using cross-view features and
treats them as positives for InfoNCE objectives.  We additionally run a baseline, denoted NNCLR (same views) in
Figure~\ref{fig:overview}, by removing the augmented view and directly maximizing the similarity between image features
and their nearest neighbor.
\item {\text SWAV}~\citep{caron2020unsupervised} maximizes view consistent objectives using clustering-based targets;
it balances the categorical assignment using sinkhorn iterations.
\item {\text DINO}~\citep{caron2021emerging} optimizes clustering-based across-views objectives via knowledge
distillation and proposes sharpening and centering techniques to prevent collapsing.
\item {\text BYOL}~\citep{grill2020bootstrap} only contains the maximizing term and adopts a momentum-updated Siamese
model to process augmented input to prevent collapsed solutions.

\end{squishlist}
In MAE~\citep{he2022masked}, we employ a VIT-small model, training it with default masking ratio and a patch-size of 4
for CIFAR experiments and 8 for ImageNet-10 experiments.

In DDPM~\citep{ho2020denoising}, we use unconditional model and train it with default hyper-parameters.  Feature are
extracted from the second decoder block with noise level at t = 11, following the optimal configurations of
\cite{xiang2023denoising}.

\subsection{Qualitative Comparison}
\label{sec:appendix_qualitative}

\begin{figure*}[!tp]
   \centering
   \includegraphics[width=\textwidth]{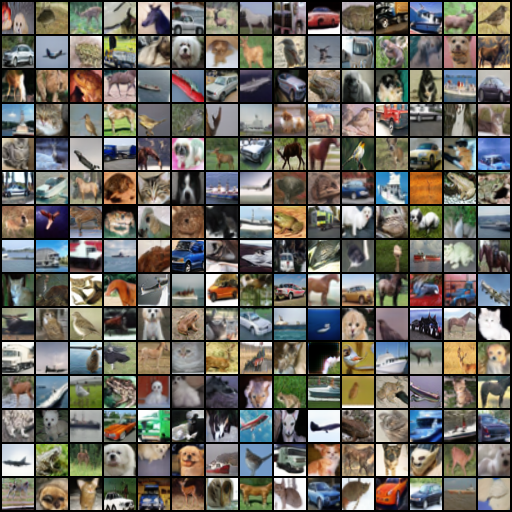}
    \caption{Randomly generated images from GAN trained with our full objectives.}
   \label{fig:sup_fig_full}
\end{figure*}

\begin{figure*}[!tp]
   \centering
   \includegraphics[width=\textwidth]{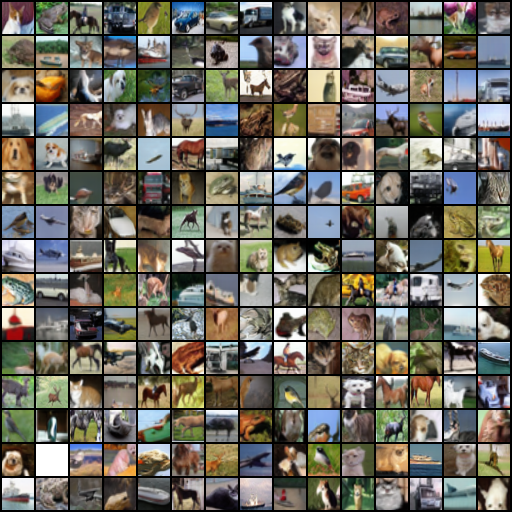}
    \caption{Randomly generated images from GAN trained with Eq.~\ref{eqn:Gaussian} only, JSD.}
   \label{fig:sup_fig_gaussian}
\end{figure*}

\begin{figure*}[!tp]
   \centering
   \includegraphics[width=\textwidth]{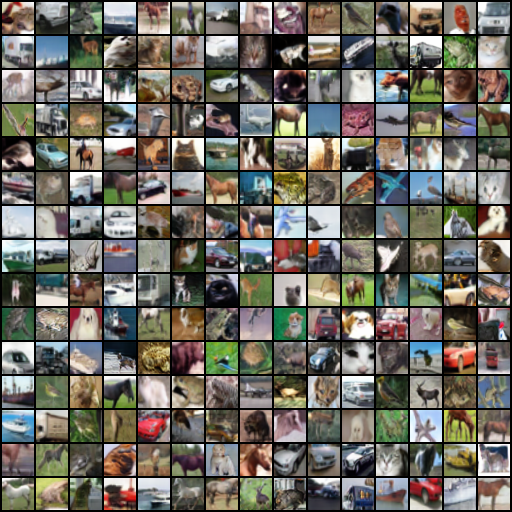}
    \caption{Randomly generated images from GAN trained with Hinge Loss.}
   \label{fig:sup_fig_hinge}
\end{figure*}

\begin{figure*}[!tp]
   \centering
   \includegraphics[width=\textwidth]{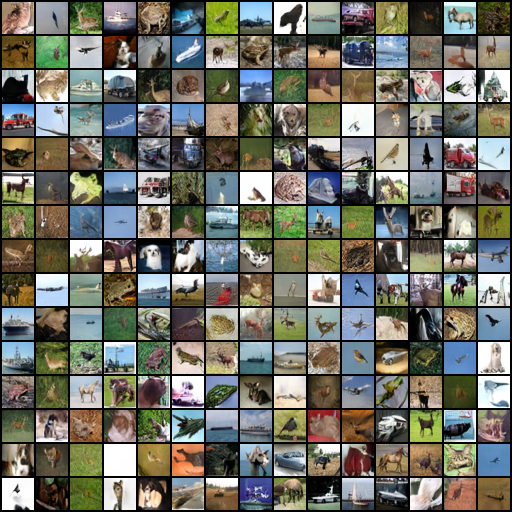}
    \caption{Randomly generated images from unconditional BigGAN~\citep{brock2018large}.}
   \label{fig:sup_fig_biggan}
\end{figure*}

We provide visualization of generated images for the following configurations:

\vspace{5pt}
\begin{squishlist}
\item Figure~\ref{fig:sup_fig_full}: Results of training with our full objectives (our method):\\
\begin{eqnarray*}
\mathcal{L^{\text{Full}}}  := \mathcal{L}_{\rm{Gaussian}} + \lambda_c\mathcal{L}_{\rm{cluster}}  +  \lambda_s \mathcal{L}_{\rm{reg}}.
\end{eqnarray*}
\item Figure~\ref{fig:sup_fig_gaussian}: Results of training with Equation~\ref{eqn:Gaussian} only, JSD: 
\begin{eqnarray*}
\mathcal{L^{\text{JSD}}}  := \mathcal{L}_{\rm{Gaussian}} + \lambda_s \mathcal{L}_{\rm{reg}}.
\end{eqnarray*}
\item Figure~\ref{fig:sup_fig_hinge}: Results of training with Hinge Loss.\\
To train with Hinge loss, we change discriminator to output a scalar: $\D(\vx)\in\mathbb{R}$ and optimize the hinge loss
defined as follows:
\\
\begin{eqnarray*}
\mathcal{L^{\text{Hinge}}} &:= & \mathcal{L}^{\text{Hinge}}_{\D} + \mathcal{L}^{\text{Hinge}}_{\G} + \lambda_s \mathcal{L}_{\text{reg}},\\ 
\mathcal{L}^{\text{Hinge}}_{\D}
&:=& \max_{\D}\left(\min\left(0, -1 + \D\left(\vx\right)\right) - \min\left(0, -1 - \D\left(\hat{\vx}\right)\right)
\right),\\
\mathcal{L}_{\G}^{\text{Hinge}}
&:=& \min_{\G}-\D(\hat{\vx}).
\end{eqnarray*}

\item Figure~\ref{fig:sup_fig_biggan}: Results of BigGAN~\citep{brock2018large}.
\end{squishlist}

\vspace{5pt}
\noindent
\textbf{Conclusion.}
We observe that training with full objectives (our method) achieves the best quality and diversity in generated images.

\clearpage
\clearpage

\end{document}